\documentclass{article}
\usepackage[a4paper,margin=1in]{geometry}
\usepackage{graphicx}
\usepackage{amsmath}
\usepackage{amssymb}
\usepackage{booktabs}
\usepackage{authblk}
\usepackage{lineno}

\title{A Neural Operator-Based Approach to Symbolic Discovery of PDEs}
\author[1]{Sergei Garmaev}
\author[1]{Olga Fink}  

\affil[1]{Intelligent Maintenance and Operations Systems Laboratory, EPFL, Lausanne, Switzerland}

\date{}

\begin{document}

\maketitle


\begin{abstract}
Discovering governing equations from data remains challenging when the underlying dynamics involve nonlocal differential operators, field interactions governed by auxiliary equations, or temporal memory effects. We propose Neural Operator-based symbolic Model approximaTion and discOvery (NOMTO), a framework that extends Equation Learner-type symbolic architectures by incorporating pretrained neural operators as nodes in the symbolic network. NOMTO represents candidate equations as sparse differentiable computational graphs that combine algebraic operations with fixed neural operator surrogates pretrained to approximate nonlinear operators. We evaluate the method on model-discovery problems involving nonlocal spatial operators, couplings mediated by auxiliary field equations, and temporal integral terms representing memory effects. The results show that NOMTO can recover compact governing equations containing nonlocal operator terms, thereby extending symbolic model discovery beyond libraries restricted to local derivatives and point-wise algebraic combinations.
\end{abstract}

\section{Introduction}

Physical systems are commonly described by mathematical equations that specify how their states evolve, interact, and respond to external forcing. Discovering such equations from observations is a longstanding goal across the natural sciences and engineering, offering a path toward interpretable models that reveal the mechanisms and structure of the observed dynamics. While many physical laws have historically been derived from first principles, increasingly complex systems are often characterized by incomplete knowledge, unresolved multiscale interactions, or phenomena for which the governing equations remain unknown. This has motivated growing interest in data-driven scientific discovery, where mathematical models are inferred directly from observations.

A major direction in this area is symbolic model discovery, which seeks to recover compact mathematical expressions that explain observed dynamics while remaining interpretable and physically meaningful. In contrast to purely predictive machine learning models, symbolic discovery aims to identify the underlying structure of the governing equations themselves, so that the recovered model can be related to known physical principles, conservation laws, constitutive assumptions, or coupling mechanisms.

Several approaches have been proposed for discovering PDEs from spatiotemporal data. PDE-FIND formulates partial differential equation (PDE) discovery as sparse regression over a prescribed library of candidate terms constructed from the field, its partial derivatives, and algebraic products of these quantities \cite{rudy2017data}. DeepMoD follows the same sparse-identification principle, but uses a neural network to approximate the observed field and to compute derivative-based library terms by automatic differentiation before selecting a sparse PDE model \cite{both2021deepmod}. Physics-informed equation discovery embeds this idea in a PINN-like framework, combining solution approximation, derivative computation, and sparse regression to identify PDEs from scarce and noisy data \cite{chen2021physics}. Another line of work uses symbolic neural-network architectures. PDE-Net 2.0 learns differential operators through convolutional filters and passes their outputs to a symbolic multi-layer network to recover the nonlinear response function of the PDE \cite{long2019pde}. The Equation Learner (EQL) architecture represents symbolic expressions as neural-network-like computational graphs with predefined symbolic operations at the nodes and trainable sparse connections between them \cite{sahoo2018learning}. Parametric EQL extends this representation to parametric systems by learning equations with shared structure and parameter-dependent coefficients \cite{zhang2023deep}. Further extensions of EQL-type models have broadened the class of symbolic structures accessible to gradient-based optimization. Complex Equation Learner (CEQL) introduces complex-valued weights to improve the recovery of rational and singular expressions involving division, logarithms, and square roots \cite{garmaev2026complex}. The same differentiable symbolic representation has also been used beyond supervised model discovery, with the Symbolic Equation Solver (SES) optimizing an EQL-type symbolic model against equation residuals and auxiliary conditions to recover explicit symbolic solutions without paired input-output data \cite{garmaev2026ses}.


Despite their methodological differences, existing PDE-discovery approaches share a common assumption: candidate equations are assembled from point-wise quantities, including field values, local derivatives, and algebraic combinations evaluated at individual space-time locations. This formulation is well suited to classical PDEs expressed through local differential operators. However, many physical systems are governed by nonlocal interactions, integral operators, or auxiliary field equations whose value depends on the state of the system over an extended spatial domain or temporal history. As a consequence, current symbolic discovery methods are fundamentally restricted to a subset of physical laws that can be expressed through local symbolic representations, leaving a broad class of nonlocal governing equations beyond their reach.

Neural operators offer a natural framework for representing such nonlocal dependencies. Rather than learning mappings between finite-dimensional vectors, neural operators learn mappings between function and can therefore approximate integral operators, history-dependent responses, and other function-level transformations.
Deep Operator Network (DeepONet) represents such mappings by decomposing the operator output into a learned basis representation, where the trunk network learns basis functions over the output coordinates and the branch network maps the input function to the corresponding coefficients in this basis \cite{lu2021learning}. Fourier Neural Operators (FNO) follow a different construction by lifting the input function to a higher-dimensional feature representation and repeatedly applying spectral convolution layers, in which the integral-kernel action is parameterized in Fourier space \cite{li2021fourier}. This makes FNO an efficient architecture for learning function-to-function maps on structured grids. Other neural-operator architectures introduce different inductive biases: multipole graph neural operators use multilevel graph representations to capture interactions across irregular discretizations \cite{li2020multipole}, Laplace Neural Operators use Laplace-domain representations for transient dynamics \cite{cao2024laplace}, and Convolutional Neural Operators (CNO) use multiscale convolutional architectures while preserving the operator-learning viewpoint \cite{raonic2024convolutional}. Overall, these methods demonstrate that neural networks can accurately approximate complex operators. However, they are primarily designed as predictive surrogates and do not directly yield interpretable governing equations.

We propose the Neural Operator-based symbolic Model approximaTion and discOvery framework (NOMTO\footnote{The name Nomto  is derived from the  Buryat-Mongolian language, translating to  "educated" or "wise", which reflects the algorithm's intelligent gradient-based exploration of the symbolic space.}), an extension of the Equation Learner (EQL) framework designed to broaden the class of governing equations that can be discovered from data. Existing EQL-based and sparse-regression-based approaches construct equations from local symbolic terms, such as field values, derivatives, and algebraic combinations. Consequently, they are fundamentally limited to governing laws that can be expressed through local symbolic libraries.

The key innovation is to treat pretrained neural operators as symbolic primitives that can participate directly in equation discovery. This transforms the symbolic search space from one containing only local differential terms to one that additionally includes nonlocal operators, integral operators, and auxiliary field mappings. Consequently, NOMTO expands the scope of symbolic discovery from classical local PDEs to a broader class of governing laws involving nonlocal physical processes.

The method represents a candidate equation as a differentiable computational graph whose edges are trainable scalar weights and whose nodes apply operations from a prescribed library. Consistent with EQL, sparsity is enforced directly on the graph coefficients, enabling compact and interpretable equations to be recovered. Unlike standard EQL architectures, however, selected nodes may be implemented as pretrained neural operators acting on discretized functions rather than point-wise evaluated quantities.

In contrast to standard EQL-type models, NOMTO allows selected nodes to be implemented as pretrained neural-operator surrogates that act on discretized functions. These nodes take discretized functions as inputs and return discretized operator outputs on the same grid, while their parameters remain fixed during symbolic discovery. During training, NOMTO minimizes the discrepancy between the graph output and the target quantity, such as a time derivative or a constitutive response, while applying a sparsity penalty to the trainable graph weights. The pretrained operator nodes remain fixed, so optimization acts only on the symbolic graph coefficients. After training, the small weights are removed and the remaining active paths define a compact equation composed of the remaining operators. 

This directly addresses a fundamental limitation of existing PDE-discovery methods, which are currently are largely restricted to local differential libraries.  By incorporating pretrained operator  nodes into the symbolic search process, NOMTO extends the discoverable model class beyond local differential equations to governing laws containing nonlocal operators, integral terms, and auxiliary solution operators. Importantly, NOMTO preserves the interpretability and sparsity that make symbolic discovery scientifically useful while extending the discoverable model class.

We evaluate NOMTO on physical systems whose governing equations cannot be represented within conventional local symbolic libraries. Fractional diffusion, requires a fractional Laplacian operator,  the Euler-Poisson system relies on an auxiliary coupling defined by a  Poisson equation, and hereditary viscoelasticity depends on an integral memory operator acting on the entire strain history.

These systems therefore provide a direct test of whether extending symbolic discovery enables the recovery of governing laws that are inaccessible to existing symbolic discovery approaches. Together, they allow us to evaluate whether neural operators can serve as reusable symbolic building blocks for scientific discovery, thereby extending symbolic regression beyond local differential equations toward a broader class of physically meaningful governing equations.

\section{Results}
We evaluate NOMTO on three discovery settings that require symbolic models to include operators acting on discretized functions. These case studies represent three distinct classes of operator-valued physics: non-local transport governed by fractional differential operators, auxiliary-field interactions arising from coupled partial differential equations, and constitutive laws with hereditary memory. The cases studies span spatial, field-mediated and temporal forms of non-locality that are inaccessible to conventional symbolic libraries. Specifically, we consider a fractional diffusion equation with a nonlocal spatial operator, an Euler-Poisson system with an auxiliary-field interaction, and a hereditary viscoelastic constitutive law with an integral memory term. For each case, we compare NOMTO with FNO, CNO, and FNN surrogates and report the discovered expressions, relative errors, recovered coefficients, and structural correctness.

\subsection{Fractional Diffusion}
As a first case, we consider the fractional diffusion equation
\begin{equation}
    u_t(x,t)+\kappa(-\Delta)^s u(x,t)=0, \qquad 0<s<1,
\end{equation}
where $u=u(x,t)$ is the unknown scalar field, $x$ is the spatial coordinate, $t$ is time, $u_t=\partial u/\partial t$, $\kappa>0$ is the diffusion coefficient, and $(-\Delta)^s$ is the fractional Laplacian. In this equation, the non-locality arises from the operator $(-\Delta)^s$, whose value at a point depends on the values of $u$ away from the point $x$, in general over the whole domain.

In all simulations, the equation parameters were fixed to $\kappa=1.0$ and $s=0.7$. The dataset consisted of $100$ simulations, each corresponding to a different initial condition. The equation was simulated on a one-dimensional periodic domain using a Fourier pseudospectral representation, for which the fractional Laplacian is diagonal in Fourier space. Across the generated dataset, only the initial condition was varied. Full details of the simulation procedure and initial-condition sampling are provided in Appendix~\ref{app:fractional_diffusion_simulation}. For compactness, we denote the fractional diffusion, Euler--Poisson, and hereditary viscoelasticity case studies by FD, EP, and HV, respectively.

\begin{table}[h]
    \centering
    \caption{Discovered symbolic models and trajectory errors. FD denotes fractional diffusion, EP denotes the Euler--Poisson system, and HV denotes hereditary viscoelasticity.}
    \label{tab:discovered_models}
    \begin{tabular}{llllcc}
        \toprule
        \textbf{Case}
        & \textbf{Target}
        & \textbf{Surrogate}
        & \textbf{Discovered expression}
        & \textbf{Rel. MSE}
        & \textbf{Correct structure} \\
        \midrule

        FD & $u_t$ & FNO & $- 1.00 \widehat{\mathcal{G}}_{L_s}^{\,\mathrm{FNO}}{\left( u \right)}$ & $9.47 \times 10^{-6}$ & Yes \\

        FD & $u_t$ & CNO & $- 0.01\,u - 0.99 \widehat{\mathcal{G}}_{L_s}^{\,\mathrm{CNO}}{\left(u\right)}$ & $3.34 \times 10^{-4}$ & Partially \\

        FD & $u_t$ & FNN & $- 1.00 \widehat{\mathcal{G}}_{L_s}^{\,\mathrm{FNN}}{\left(u \right)}$ & $1.29 \times 10^{-2}$ & Yes \\

        \midrule

        EP & $\rho_t$ & FNO & $- 0.25 \rho \rho_x - 0.97 \rho u_x - 0.97 u \rho_x $ & $1.288e-02$ & Partially \\

         &  &  & $- 0.08 u u_x + 0.25 \rho_x - 0.05 u_x$ &  &  \\
        
        EP
        & $\rho_t$
        & CNO
        & $- 1.0 \rho u_x - 1.0 u \rho_x$
        & $2.653e-03$
        & Yes \\

        EP
        & $\rho_t$
        & FNN
        & $8.7 \rho - 8.7$
        & $9.337e-01$
        & No \\

        EP
        & $u_t$
        & FNO
        & $1.0 \widehat{\mathcal{G}}_{P}^{\,\mathrm{FNO}}{\left(\rho \right)} - 0.97 u u_x$
        & $1.150e-03$
        & Yes \\

        EP
        & $u_t$
        & CNO
        & $1.0 \widehat{\mathcal{G}}_{P}^{\,\mathrm{CNO}}{\left(\rho \right)} - 1.0 u u_x$
        & $3.245e-04$
        & Yes \\

        EP
        & $u_t$
        & FNN
        & $1.0 \widehat{\mathcal{G}}_{P}^{\,\mathrm{FNN}}{\left(\rho \right)} - 1.0 u u_x$
        & $1.856e-03$
        & Yes \\

        \midrule

        HV
        & $\sigma$
        & FNO
        & $0.5 \widehat{\mathcal{G}}_{H}^{\,\mathrm{FNO}}{\left(\varepsilon \right)} + 1.0 \varepsilon$
        & $5.384e-06$
        & Yes \\

        HV
        & $\sigma$
        & CNO
        & $0.5 \widehat{\mathcal{G}}_{H}^{\,\mathrm{CNO}}{\left(\varepsilon \right)} + 1.0 \varepsilon$
        & $3.762e-06$
        & Yes \\

        HV
        & $\sigma$
        & FNN
        & $0.49 \widehat{\mathcal{G}}_{H}^{\,\mathrm{FNN}}{\left(\varepsilon \right)} + 1.0 \varepsilon$
        & $3.903e-06$
        & Yes \\

        \bottomrule
    \end{tabular}
\end{table}

The results for all case studies are summarized in Tables~\ref{tab:discovered_models} and~\ref{tab:recovered_coefficients}. For the fractional diffusion case, the task of NOMTO is to recover the expected single-operator structure with both the FNO and FNN surrogates:
\begin{equation}
    u_t \approx -\widehat{\mathcal{G}}_{L_s}^{\,a}[u],
\end{equation}
where $a$ denotes the surrogate architecture. The FNO-based model gives the most accurate result, with relative MSE $9.47\times 10^{-6}$ and recovered coefficient $\hat{\kappa}=1.00$. The FNN-based model also recovers the correct symbolic structure and coefficient, but its trajectory error is higher, with relative MSE $1.29\times 10^{-2}$.

The CNO-based model recovers the dominant fractional-Laplacian term with coefficient $0.99$, but also retains a small additional linear term, $-0.01u$. We therefore classify its structure as partially correct. Its relative MSE, $3.34\times 10^{-4}$, remains low. Overall, the fractional diffusion experiment demonstrates that NOMTO can recover governing equations whose dominant physical mechanism is a genuinely non-local  operator. Despite the absence of an analytical fractional-Laplacian primitive in the symbolic library, the correct operator structure and coefficient are recovered with high accuracy. These results indicate that pretrained neural operators can act as faithful symbolic proxies for non-local physical mechanisms, enabling symbolic discovery beyond the class of equations expressible through local differential operators alone.

\begin{table}[h]
    \centering
    \caption{Recovered coefficients in the discovered symbolic models. FD denotes fractional diffusion, EP denotes the Euler--Poisson system, and HV denotes hereditary viscoelasticity.}
    \label{tab:recovered_coefficients}
    \begin{tabular}{llcccc}
        \toprule
        \textbf{Case}
        & \textbf{Coefficient}
        & \textbf{True value}
        & \textbf{FNO}
        & \textbf{CNO}
        & \textbf{FNN} \\
        \midrule

        FD
        & $\kappa$
        & $1.0$
        & $1.00$
        & $0.99$
        & $1.00$ \\

        EP
        & coefficient of $u u_x$
        & $-1.0$
        & $-0.97$
        & $-1.00$
        & $-1.00$ \\

        EP
        & coefficient of Poisson force
        & $1.0$
        & $1.00$
        & $1.00$
        & $1.00$ \\

        HV
        & $E$
        & $1.0$
        & $1.00$
        & $1.00$
        & $1.00$ \\

        HV
        & $A$
        & $0.5$
        & $0.50$
        & $0.50$
        & $0.49$ \\

        \bottomrule
    \end{tabular}
\end{table}

\subsection{Euler-Poisson System}
As a second case, we consider the one-dimensional Euler--Poisson system
\begin{equation}
    \rho_t + (\rho u)_x = 0, \qquad
    u_t + u\,u_x = -\phi_x, \qquad
    -\phi_{xx}=\rho-1,
\end{equation}
where $\rho=\rho(x,t)$ is the density, $u=u(x,t)$ is the velocity, $\phi=\phi(x,t)$ is the self-consistent potential, $x$ is the spatial coordinate, and $t$ is time. Here, the force term $-\phi_x$ is non-local because it is obtained by solving the auxiliary Poisson equation with $\rho$ as the source term. Thus, the acceleration at a point depends on the spatial distribution of the density over the whole domain, rather than only on local values of $\rho$ and its derivatives.

The dataset consisted of $100$ simulations, each corresponding to a different smooth initial condition. The system was simulated on the same one-dimensional periodic spatial grid as the fractional diffusion case, with $L=2\pi$ and $N_x=256$. The Poisson equation was solved spectrally at each time step, and the hyperbolic evolution equations for $\rho$ and $u$ were advanced using a fourth-order Runge--Kutta method. To keep the solutions smooth while making the nonlinear and non-local terms sufficiently expressed in the data, the simulations used moderate-amplitude perturbations of the uniform-density state and were integrated over the short time interval $t\in[0,0.5]$. In the symbolic discovery stage, the density perturbation $r=\rho-1$ and the velocity $u$ were used as input fields. Full simulation details are provided in Appendix~\ref{app:euler_poisson_simulation}.

The Euler--Poisson system presents a more demanding discovery problem because it combines local nonlinear transport with a non-local interaction mediated through the solution of an auxiliary Poisson equation. Recovering the governing equations therefore requires identifying both conventional differential terms and a field coupling within a single symbolic model. For the continuity equation, the expected structure is
\begin{equation}
    \rho_t = -(\rho u)_x = -\rho u_x - u\rho_x .
\end{equation}
As shown in Table~\ref{tab:discovered_models}, the CNO-based model recovers this structure directly, with relative MSE $2.653\times10^{-3}$. The FNO-based model recovers the dominant transport terms $-\rho u_x-u\rho_x$ with coefficients close to $-1$, but also retains additional derivative terms involving $\rho\rho_x$, $u u_x$, $\rho_x$, and $u_x$. We therefore classify the FNO continuity model as partially correct. The FNN model does not recover the continuity structure and instead returns a pointwise density-dependent expression, leading to substantially larger error.

For the momentum equation, the target structure is
\begin{equation}
    u_t = -u u_x - \phi_x ,
\end{equation}
where the force term is represented in NOMTO through the Poisson-gradient surrogate. In the notation of the discovered models, this corresponds to a combination of the nonlinear convective term $u u_x$ and the surrogate operator $\widehat{\mathcal{G}}_{P}^{\,a}[\rho]$. All three surrogate architectures recover this structure, as reported in Table~\ref{tab:discovered_models}. The recovered coefficients are also close to the true values (Table~\ref{tab:recovered_coefficients}), with convective-term   coefficients of $-0.97$, $-1.00$, and $-1.00$, and a Poisson-force coefficient of $1.00$  for all three surrogate architectures.

The lowest momentum-equation error is obtained with the CNO surrogate, with relative MSE $3.245\times10^{-4}$. The FNO and FNN variants also recover the correct symbolic structure, with relative MSE values $1.150\times10^{-3}$ and $1.856\times10^{-3}$, respectively. 

Overall, these results demonstrate that NOMTO can recover governing laws that combine local and non-local mechanisms. In contrast to the fractional diffusion example, where the dominant physical process is represented by a single non-local operator, in the Euler--Poisson system the auxiliary-field coupling does not need to be solved explicitly during symbolic discovery, because the pretrained Poisson-gradient surrogate provides the force field directly from the density perturbation. The successful recovery of this term together with the nonlinear transport term shows that pretrained neural-operator surrogates can make auxiliary-field couplings available as reusable symbolic primitives in coupled physical systems.

\subsection{Hereditary Viscoelasticity}
As a fourth case, we consider the hereditary viscoelastic constitutive law
\begin{equation}
    \sigma(t)=E\,\varepsilon(t)+\int_0^t A e^{-(t-s)/\tau}\,\varepsilon(s)\,ds,
\end{equation}
where $\sigma=\sigma(t)$ is the stress, $\varepsilon=\varepsilon(t)$ is the strain, $E>0$ is the instantaneous elastic modulus, $A>0$ is the memory amplitude, $\tau>0$ is the relaxation time, and $s$ is the past-time integration variable. In contrast to the previous cases, the non-locality is temporal rather than spatial: the stress at time $t$ depends on the full past strain history through the hereditary integral.

This example differs fundamentally from the previous case studies because the non-locality is temporal rather than spatial. The constitutive response at time (t) depends on the entire past deformation history, making it a representative example of a governing law with a temporal memory operator.

The dataset consisted of $100$ simulations, each corresponding to a different smooth strain history. The parameters of the constitutive law were fixed to $E=1.0$, $A=0.5$, and $\tau=0.1$ for all simulations. The strain histories were generated on the temporal interval $t\in[0,1]$ using $N_t=256$ equidistant time points. Across the generated dataset, only the input strain history was varied. The hereditary integral was evaluated through an equivalent internal memory variable, which provides a causal and stable way to compute the stress history. Full simulation details are provided in Appendix~\ref{app:hereditary_viscoelasticity_simulation}.

The expected symbolic structure is
\begin{equation}
    \sigma(t) = E\varepsilon(t) + A\mathcal{H}[\varepsilon](t),
\end{equation}
where $\mathcal{H}$ denotes the hereditary memory operator. In NOMTO, this operator is represented by the pretrained surrogate $\widehat{\mathcal{G}}_{H}^{\,a}$, with $a$ denoting the surrogate architecture.

As shown in Table~\ref{tab:discovered_models}, all three surrogate architectures recover the correct elastic-plus-memory structure. The FNO and CNO variants recover
\begin{equation}
    \sigma \approx 1.0\varepsilon + 0.5\widehat{\mathcal{G}}_{H}^{\,a}[\varepsilon],
\end{equation}
matching the true coefficients $E=1.0$ and $A=0.5$. The FNN variant gives the same structure with a slightly lower memory coefficient, $\hat{A}=0.49$. The recovered coefficients are summarized in Table~\ref{tab:recovered_coefficients}.

The trajectory errors are small for all three variants. The relative MSE values are $5.384\times10^{-6}$ for FNO, $3.762\times10^{-6}$ for CNO, and $3.903\times10^{-6}$ for FNN. Thus, in this case, NOMTO consistently identifies the instantaneous elastic contribution and the temporal memory contribution. Unlike the previous examples, the hereditary constitutive law depends on the complete temporal history of the system rather than on spatial interactions. The successful recovery of both the elastic and memory contributions therefore demonstrates that NOMTO is not restricted to spatial operators, but can discover governing laws containing general history-dependent functionals. This substantially broadens the class of interpretable models that can be recovered using symbolic discovery.

The three case studies demonstrate that NOMTO expands symbolic discovery beyond equations that contain local differential operators. The recovered models contain three distinct classes of operator-valued mechanisms: non-local spatial operators in fractional diffusion, auxiliary-field mappings in the Euler--Poisson system, and temporal memory operators in hereditary viscoelasticity. Despite their fundamentally different physical interpretations, all three can be represented within a common symbolic framework through pretrained neural operator surrogates.

These results indicate that symbolic discovery can be extended from local symbolic expressions built from point-wise quantities to governing equations that explicitly contain nonlocal operators. By embedding learned operators as symbolic primitives, NOMTO substantially enlarges the class of scientific models that can be recovered from data while retaining the interpretability and sparsity that make symbolic approaches attractive for scientific discovery.

\section{Discussion} \label{sec:discussion}

The three case studies demonstrate that NOMTO expands symbolic discovery beyond local analytical libraries. The recovered models contain three distinct classes of operator-valued mechanisms: non-local spatial operators in fractional diffusion, auxiliary-field mappings in the Euler--Poisson system, and temporal memory operators in hereditary viscoelasticity. Despite their fundamentally different physical interpretations, all three can be represented within a common symbolic framework through pretrained neural operator surrogates.

These results suggest a shift from symbolic discovery of functions to symbolic discovery of operator-valued physical laws. Existing symbolic discovery approaches are typically restricted to combinations of local analytical functions and differential operators. Consequently, many governing equations that depend on non-local interactions, auxiliary fields or accumulated history lie outside the representational scope of conventional symbolic libraries. By embedding learned operators as symbolic primitives, NOMTO substantially enlarges the class of scientific models that can be recovered from data while retaining the interpretability and sparsity that make symbolic approaches attractive for scientific discovery.

By using pretrained neural-operator surrogates as fixed differentiable nodes in the symbolic graph, NOMTO can represent nonlocal spatial operators, auxiliary-field interactions, and temporal memory effects within a sparse symbolic architecture. Since these surrogate nodes are differentiable with respect to their inputs, gradients can be propagated through the operator evaluations during symbolic optimization. This allows NOMTO to learn not only the external coefficients multiplying operator terms, but also the trainable combinations of variables that appear inside the operator arguments.

A key feature of NOMTO is the separation of operator learning from equation discovery. Neural operators are trained independently to approximate physically meaningful operator actions and are subsequently frozen during symbolic optimization. This transforms the recovery of operator-valued governing equations into a sparse structure-identification problem, allowing the discovery stage to focus on identifying which operators participate in the governing law and how they interact. The resulting modular formulation enables operator libraries to be reused across multiple discovery tasks and provides a natural mechanism for incorporating prior physical knowledge through the choice of candidate operators.

At the same time, the proposed framework inherits several limitations. First, the quality of the recovered symbolic model depends on the fidelity of the surrogate operators. Because the discovered equations contain learned operator approximations rather than exact analytical operators, surrogate errors can propagate into both the recovered coefficients and the extracted symbolic structure. Improvements in neural operator architectures, training strategies and uncertainty quantification are therefore expected to directly improve symbolic recovery performance.

Second, NOMTO assumes that the relevant classes of physical operators are represented within the candidate library. Although the complete governing equation need not be known in advance, the construction of the operator library still requires some degree of prior knowledge regarding plausible physical mechanisms. Developing methods for adaptive operator-library construction or automatic operator discovery represents an important direction for future research.

At the same time, the operators used in NOMTO are learned approximations rather than exact analytical operations. Their errors can therefore propagate into the NOMTO approximation and affect the extracted symbolic expression. More accurate and robust neural-operator surrogates in future are expected to improve the accuracy of the discovered models and their internal coefficients.
More broadly, NOMTO illustrates how advances in operator learning can be integrated with interpretable scientific machine learning. Rather than treating neural operators as end-to-end predictive models, the proposed framework uses them as reusable components within an interpretable symbolic representation. This perspective may enable symbolic discovery in a wide range of scientific domains where governing mechanisms are naturally expressed through function-space operators rather than local analytical expressions.

\section{Methods} \label{sec:methods}

\subsection{Overview of NOMTO}

The proposed NOMTO framework extends the EQL architecture to discover governing equations containing non-local operators. In the original EQL formulation, symbolic expressions are represented as differentiable computational graphs whose nodes apply predefined symbolic operations and whose trainable connections determine the structure of the recovered equation. While highly effective for discovering equations composed of local quantities and algebraic transformations, EQL is inherently limited by the operation library available to the symbolic graph. NOMTO addresses this limitation by augmenting the EQL operation library with pretrained neural-operator surrogates that approximate both local and nonlocal physical operators.

The graph combines input variables and intermediate quantities through trainable weighted connections, while its nodes apply operations from a predefined set of operations. The resulting operation library contains both elementary functions and operators acting on discretized functions. Some nodes are therefore implemented as pretrained neural-operator surrogates, which approximate the required operators, including partial derivatives, fractional Laplacian operators, Poisson-type solution operators, and hereditary integral operators. After pretraining, these surrogate operators remain fixed and symbolic discovery proceeds by optimizing only the graph coefficients. This separation of operator learning and equation discovery enables NOMTO to identify sparse governing equations containing both local and nonlocal operators.

The following subsections describe the NOMTO architecture, pretraining of the surrogate models, the optimization and expression extraction procedure.

\subsection{Symbolic Model Architecture} \label{sec:architecture_overview}
A fundamental limitation of existing symbolic discovery methods is that they operate within libraries of local analytical functions. Consequently, they can recover governing equations composed of algebraic terms and local differential operators, but are not naturally suited for representing physical laws whose essential terms encode non-local spatial interactions or temporal history dependence. These mechanisms are common in physical systems, where the evolution at a given point or time may depend on spatially distributed fields, auxiliary equations, or past states rather than only on point-wise state variables. These operators cannot be naturally represented within the local symbolic libraries employed by existing symbolic discovery frameworks. NOMTO overcomes this limitation by introducing neural operator surrogates as symbolic primitives. Rather than restricting the search space to local analytical expressions, NOMTO augments symbolic computation graphs with pretrained neural approximations of physically meaningful operators.

A central design principle of NOMTO is the separation of operator learning from equation discovery. Neural operators are first trained independently to approximate target physical operators and are subsequently frozen. During symbolic discovery, these surrogates act as differentiable operator-valued building blocks that can be combined with conventional algebraic terms within a single symbolic graph. In this way, NOMTO preserves the interpretability and sparsity of symbolic regression while substantially expanding the class of physical laws that can be represented.

\begin{figure}[h]
    \centering
    \includegraphics[width=0.9\linewidth]{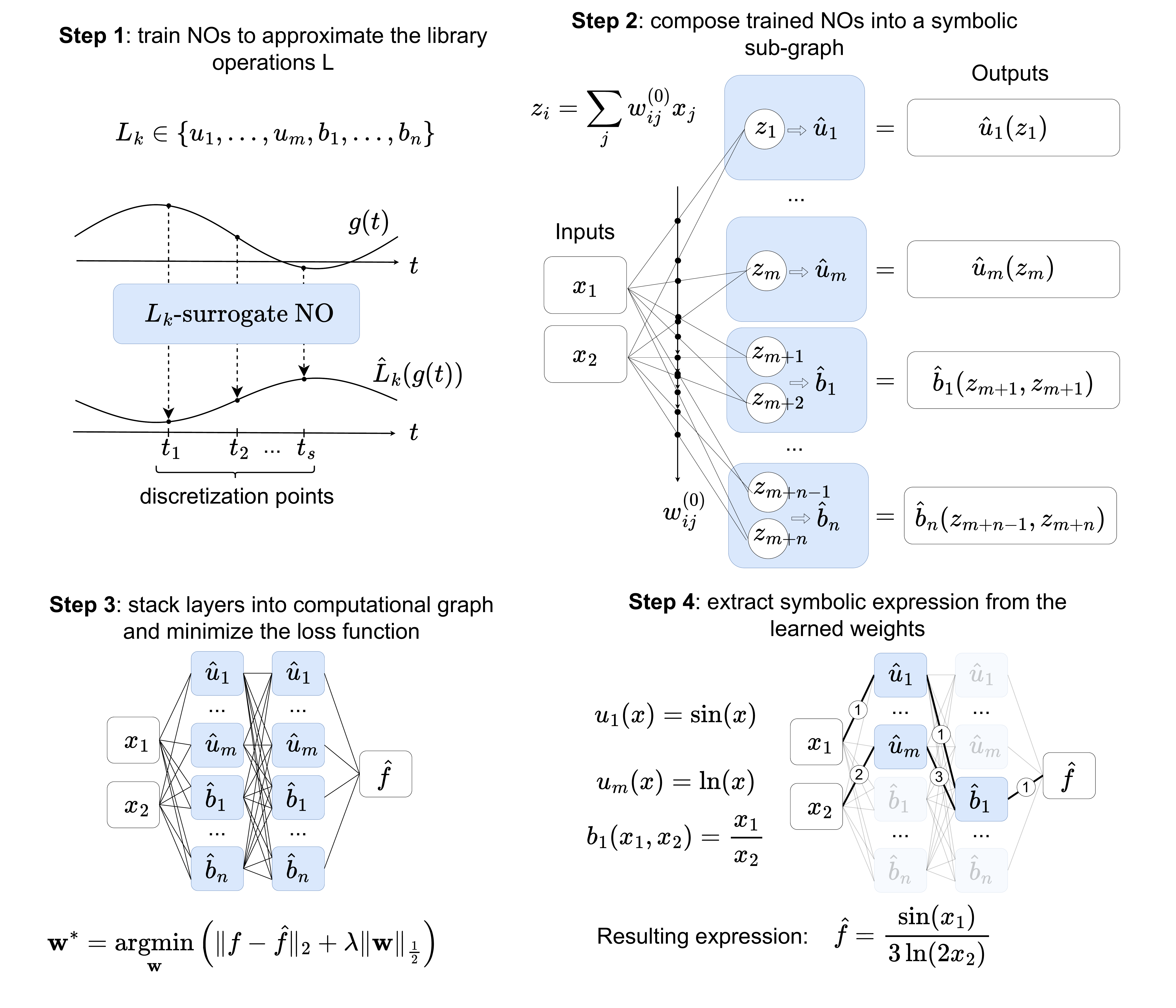}
    \caption{General architecture of the Neural Operator-based symbolic Model approximaTion and discOvery (NOMTO) algorithm. NOMTO uses an EQL-based symbolic graph in which selected operation nodes are implemented as pretrained neural-operator surrogates.}
    \label{fig:nomto_architecture}
\end{figure}

The architecture builds upon the layered symbolic computational graph introduced in Equation Learner (EQL) but generalizes its operation library beyond analytical functions. Selected operation nodes are replaced by pretrained neural-operator surrogates, enabling local algebraic transformations and non-local operator evaluations to participate in the same symbolic composition process. As in EQL, the symbolic expression is represented as a layered computational graph. These surrogate nodes participate in symbolic composition in exactly the same manner as standard EQL operations, enabling the discovery process to combine local algebraic terms and function-level operators within a unified symbolic graph. Each node receives a weighted sum of outputs from the previous layer and then applies a node operation. By stacking such layers, the graph builds nonlinear combinations of the input quantities. The overall architecture is shown in Fig.~\ref{fig:nomto_architecture}.

Let $\mathbf{x}^{(l-1)}$ denote the vector of node outputs from layer $l-1$. For each node in layer $l$, NOMTO first constructs a trainable linear combination
\begin{equation}
    z_i^{(l)} = \sum_j w_{ij}^{(l-1)} x_j^{(l-1)},
\end{equation}
where $x_j^{(l-1)}$ is the output of node $j$ in the previous layer, $w_{ij}^{(l-1)}$ is the trainable weight connecting node $j$ to the intermediate variable $z_i^{(l)}$, and $z_i^{(l)}$ is the input passed to the operation at node $i$. For unary operations, the node output is
\begin{equation}
    x_i^{(l)} = \hat{u}_i(z_i^{(l)}),
\end{equation}
where $\hat{u}_i$ is the unary surrogate operation assigned to node $i$.

For binary operations, two independent weighted sums are created,
\begin{equation}
    z_{i,1}^{(l)} = \sum_j w_{i1,j}^{(l-1)} x_j^{(l-1)},
    \qquad
    z_{i,2}^{(l)} = \sum_j w_{i2,j}^{(l-1)} x_j^{(l-1)},
\end{equation}
and the node output is
\begin{equation}
    x_i^{(l)} = \hat{b}_i(z_{i,1}^{(l)},z_{i,2}^{(l)}).
\end{equation}
Thus, every operation node has its own trainable multipliers to the input quantities.

The operation library contains algebraic operations and pretrained surrogate operators. Algebraic operations, such as multiplication and addition through weighted sums, are evaluated directly. Derivative and nonlocal operators are represented by neural-operator surrogates trained before the NOMTO optimization. These surrogate nodes take discretized functions as inputs and return discretized operator outputs on the same grid. Their parameters remain fixed during the symbolic discovery stage.

The final layer is connected to an output node that forms a trainable weighted combination of the generated terms. This output represents the discovered right-hand side of the equation or target quantity. After optimization, weights with negligible contribution are removed, and the remaining active graph is translated into an explicit symbolic equation.

\subsection{Neural Operator Surrogate Models} \label{sec:surrogate_models}

The central idea of NOMTO is to represent physically meaningful operators through learned approximations that can be embedded directly into symbolic expressions. Unlike conventional neural networks, which learn mappings between finite-dimensional vectors, neural operators learn mappings between function spaces and therefore provide a natural framework for approximating differential, integral and auxiliary solution operators \cite{kovachki2023neural,zappala2024learning}. This capability allows operator-valued physical mechanisms to be incorporated into symbolic discovery as reusable computational primitives.

Let $\mathcal{G}_m$ denote an operator of type $m$, where $m$ identifies the operation being approximated. In this work, the relevant operator types include first and second spatial derivatives, the fractional Laplacian, the Poisson-gradient map, and the hereditary memory integral. For each operator type $m$ and surrogate architecture $a$, we train a fixed approximation
\begin{equation}
    \widehat{\mathcal{G}}_{m}^{\,a}: u \mapsto \widehat{\mathcal{G}}_{m}^{\,a}[u],
\end{equation}
where $u$ is a discretized input function and $\widehat{\mathcal{G}}_{m}^{\,a}[u]$ is the corresponding discretized surrogate output. The superscript $a\in\{\mathrm{FNO},\mathrm{CNO},\mathrm{FNN}\}$ denotes the surrogate architecture. During symbolic discovery, the parameters of $\widehat{\mathcal{G}}_{m}^{\,a}$ are fixed and only the scalar weights of the NOMTO symbolic graph are optimized.

This notation distinguishes the exact reference operator $\mathcal{G}_m$ from its learned approximation $\widehat{\mathcal{G}}_{m}^{\,a}$. For example, $\widehat{\mathcal{G}}_{L_s}^{\,\mathrm{FNO}}$ denotes the FNO surrogate for the fractional Laplacian, $\widehat{\mathcal{G}}_{P}^{\,\mathrm{FNO}}$ denotes the FNO surrogate for the Poisson-gradient map, and $\widehat{\mathcal{G}}_{H}^{\,\mathrm{FNO}}$ denotes the FNO surrogate for the hereditary memory integral. The discovered symbolic expressions therefore contain operators of the form $\widehat{\mathcal{G}}_{m}^{\,a}$ rather than the exact operators $\mathcal{G}_m$.

The surrogate models play a fundamentally different role from conventional machine-learning models used for system identification. Their purpose is not to predict system evolution directly, but to provide accurate approximations of operator actions that can subsequently be reused during symbolic discovery. The resulting framework is modular: operator approximation and equation discovery are treated as separate learning problems. We use FNOs, which parameterize part of the operator action through spectral convolution layers \cite{li2021fourier}, and CNOs, which use multiscale convolutional architectures \cite{raonic2024convolutional}. FNOs are particularly suitable for the periodic spatial operators used in the fractional diffusion and Euler--Poisson cases, while CNOs provide an alternative multiresolution operator-learning model. We also train a fully connected neural network (FNN) baseline by flattening the discretized input and output functions. Unlike FNOs and CNOs, the FNN does not explicitly exploit the functional structure of the data.

For each operator type $m$, the surrogate training data are generated independently of the downstream simulation trajectories. Consequently, the surrogate operators learn generic operator behaviour rather than information specific to any particular governing equation. This separation prevents information leakage between operator learning and equation identification and enables operator libraries to be transferred across different discovery problems.

We sample smooth input functions $u_i$ as Gaussian mixtures on the grid used in the corresponding case study and compute the target output by applying the reference numerical operator,
\begin{equation}
    y_i^{(m)} = \mathcal{G}_m[u_i].
\end{equation}
For spatial operators, the targets are computed on the periodic domain $[0,2\pi]$ with $256$ grid points. Spatial derivatives, the fractional Laplacian, and the Poisson-gradient map are evaluated using pseudospectral method. For the hereditary case, the target is the causal memory response on the temporal grid of the viscoelasticity simulations, computed using the reference exponential-memory update. The surrogate models are then trained on max-absolute-normalized input-output pairs,
\begin{equation}
    \widetilde{u}_i = \frac{u_i}{\|u_i\|_{\infty}},
    \qquad
    \widetilde{y}_i^{(m)} =
    \frac{y_i^{(m)}}{\|y_i^{(m)}\|_{\infty}},
\end{equation}
by minimizing the mean squared error between $\widehat{\mathcal{G}}_{m}^{\,a}[\widetilde{u}_i]$ and $\widetilde{y}_i^{(m)}$.

\begin{table}[h]
    \centering
    \caption{Test MSE of pretrained surrogate models on normalized data.}
    \label{tab:no_surrogate_mse}
    \begin{tabular}{lccc}
        \toprule
        \textbf{Operation} & \textbf{FNO} & \textbf{CNO} & \textbf{FNN} \\
        \midrule
        Fractional Laplacian
        & $8.25 \times 10^{-6}$
        & $6.54 \times 10^{-4}$
        & $2.13 \times 10^{-3}$ \\

        Poisson-to-field operator
        & $1.12 \times 10^{-5}$
        & $3.17 \times 10^{-4}$
        & $7.23 \times 10^{-1}$ \\

        Hereditary integral operator
        & $1.67 \times 10^{-5}$
        & $1.36 \times 10^{-5}$
        & $7.34 \times 10^{-5}$ \\
        \bottomrule
    \end{tabular}
\end{table}

For every operator type and architecture, we use $4096$ training samples, $512$ validation samples, and $512$ test samples. Training is performed with mini-batches of size $64$. The checkpoint with the lowest validation error is retained and evaluated on the held-out test set. Table~\ref{tab:no_surrogate_mse} reports the resulting test errors for each pretrained surrogate.

In all case studies, the NOMTO symbolic graph used two symbolic layers. For the fractional diffusion and Euler--Poisson cases, the same spatial operation library was used in both layers. This library consisted of the identity operation $\mathrm{id}$, multiplication $\times$, squaring $(\cdot)^2$, the first-derivative surrogate $\widehat{\mathcal{G}}_{\partial_x}^{\,a}$, the second-derivative surrogate $\widehat{\mathcal{G}}_{\partial_{xx}}^{\,a}$, the fractional-Laplacian surrogate $\widehat{\mathcal{G}}_{L_s}^{\,a}$, and the Poisson-gradient surrogate $\widehat{\mathcal{G}}_{P}^{\,a}$. For the hereditary viscoelasticity case, the same two-layer symbolic graph was used, but with a temporal operation library consisting of the identity operation $\mathrm{id}$, multiplication $\times$, squaring $(\cdot)^2$, and the hereditary integral surrogate $\widehat{\mathcal{G}}_{H}^{\,a}$. Thus, the first two case studies used the same spatial search space, whereas the viscoelasticity case used a library adapted to temporal memory effects.

\subsection{Optimization and Expression Extraction} \label{sec:optimization_extraction}

A central design principle of NOMTO is the separation of operator learning from equation discovery. Neural operators are trained independently to approximate the desired physical operators and are subsequently frozen. Symbolic discovery is then performed solely through optimization of the graph coefficients, allowing the search process to focus on identifying the governing equation structure rather than relearning operator behaviour. By freezing the operator surrogates prior to symbolic discovery, NOMTO transforms the recovery of operator-valued governing equations into a sparse structure-identification problem. The discovery stage therefore focuses on determining which operators participate in the governing law and how they interact, rather than simultaneously learning both operator behaviour and equation structure. For a given input field vector $\mathbf{x}$ and target quantity $y$, such as a time derivative or a constitutive response, the symbolic graph defines the prediction
\begin{equation}
    \widehat{y}=\mathcal{F}_{\theta}(\mathbf{x}),
\end{equation}
where $\mathcal{F}_{\theta}$ is the NOMTO graph and $\theta$ denotes all trainable graph weights. The data-fitting loss is the mean squared error
\begin{equation}
    \mathcal{L}_{\mathrm{data}}(\theta)
    =
    \frac{1}{N}\sum_{i=1}^{N}
    \left\|
        \mathcal{F}_{\theta}(\mathbf{x}_i)-y_i
    \right\|_2^2,
\end{equation}
where $N$ is the number of training samples, $\mathbf{x}_i$ is the input sample, and $y_i$ is the corresponding reference target.

The optimization strategy follows a coarse-to-sparse paradigm. An initially dense symbolic graph is first allowed to explore candidate interactions, after which sparsity constraints progressively eliminate low-contribution pathways and a final refitting stage refines the surviving coefficients. The optimization is performed in three phases, each run for $1000$ epochs. In the first phase, the full symbolic graph is trained without sparsity regularization. This phase allows the initially dense graph to fit the target dynamics and to assign useful weights to candidate paths before any pruning is applied. The objective is therefore
\begin{equation}
    \mathcal{L}^{(1)}(\theta)=\mathcal{L}_{\mathrm{data}}(\theta).
\end{equation}

In the second phase, an $\ell_1$ penalty is added to promote sparse graph connectivity,
\begin{equation}
    \mathcal{L}^{(2)}(\theta)
    =
    \mathcal{L}_{\mathrm{data}}(\theta)
    +
    \lambda_1
    \sum_{w\in\theta_{\mathrm{reg}}}|w|,
\end{equation}
where $\lambda_1$ is the sparsity coefficient and $\theta_{\mathrm{reg}}$ is the set of graph weights used for regularization. We used $\lambda_1=10^{-4}$ for all experiments. During this phase, pruning is applied periodically. At each pruning step, active graph weights are ranked by magnitude, and a prescribed fraction of the smallest active weights is removed by setting both their values and masks to zero. A minimum number of active edges is retained to avoid collapsing the graph prematurely. This phase therefore combines sparsity regularization with discrete weight magnitude-based pruning.

In the third phase, the pruned symbolic graph is trained again without the $\ell_1$ penalty,
\begin{equation}
    \mathcal{L}^{(3)}(\theta)=\mathcal{L}_{\mathrm{data}}(\theta).
\end{equation}
This final refitting phase adjusts the remaining coefficients after sparsification. Since the inactive edges remain masked, optimization is restricted to the remaining graph structure. This separates structure selection, mainly performed during the second phase, from coefficient refinement, performed during the third phase.

All three phases use gradient-based optimization with Adam \cite{kingma2015adam} and a learning rate of $10^{-2}$. Since the exact operations and pretrained neural-operator surrogates are differentiable with respect to their inputs, gradients propagate through the entire symbolic graph. Consequently, NOMTO can optimize not only the external coefficients multiplying candidate terms, but also the trainable linear combinations that form the arguments of operator nodes.

A distinguishing feature of NOMTO is that the final model remains interpretable despite containing neural operator components. Because surrogate operators occupy explicit positions within the symbolic graph, the discovered model can be translated directly into a symbolic representation in which operator-valued terms appear alongside conventional analytical expressions. After the three training phases, an additional small-magnitude threshold is applied to remove residual negligible weights. The resulting pruned graph is then converted into a symbolic expression. Expression extraction is performed by traversing the active computational graph from the input nodes to the output node. Active weighted sums are converted into scalar coefficients, exact operation nodes are written as their corresponding algebraic or differential operations, and neural-operator surrogate nodes are written using the notation introduced in Section~\ref{sec:surrogate_models}. For example, the extracted expressions may contain $\widehat{\mathcal{G}}_{L_s}^{\,a}$ for the fractional-Laplacian surrogate, $\widehat{\mathcal{G}}_{P}^{\,a}$ for the Poisson-gradient surrogate, or $\widehat{\mathcal{G}}_{H}^{\,a}$ for the hereditary memory surrogate, where $a$ denotes the surrogate architecture.

The final output is a sparse and interpretable governing equation composed of the original input fields, active symbolic graph coefficients, exact operations, and pretrained surrogate operators. Unlike conventional black-box neural models, the resulting representation explicitly reveals both the algebraic structure of the governing law and the role of non-local operator-valued mechanisms. NOMTO therefore preserves interpretability while extending symbolic discovery to classes of physical systems that cannot be represented using local symbolic libraries alone. The recovered expression is then evaluated by comparing its predictions against the reference target and by checking whether the extracted structure and coefficients match the expected governing equation.

\section{Conclusion} \label{sec:conclusion}
We introduced NOMTO, a symbolic model-discovery framework that extends EQL-type architectures from local analytical libraries to operator-valued representations. By embedding pretrained neural operators as symbolic primitives within a sparse symbolic graph, NOMTO enables the recovery of governing laws containing non-local spatial operators, auxiliary-field interactions and temporal memory effects. This makes it possible to search for sparse equations that combine algebraic terms with nonlocal spatial operators, auxiliary-field interactions, and temporal memory operators, including integral terms. Across the fractional diffusion, Euler-Poisson, and hereditary viscoelasticity case studies, NOMTO successfully recovered compact symbolic models with the expected operator structure  and accurately identified the dominant physical coefficients. These examples demonstrate that pretrained neural operators can serve as effective symbolic proxies for physically meaningful operators, allowing governing equations to be discovered even when key mechanisms cannot be represented through conventional symbolic primitives.

The proposed framework establishes a connection between neural operator learning and symbolic scientific discovery. By separating operator approximation from equation identification, NOMTO preserves interpretability while substantially enlarging the class of physical systems that can be addressed through symbolic discovery. More broadly, the results suggest a transition from symbolic discovery of functions to symbolic discovery of operator-valued physical laws.

Future work will focus on improving surrogate fidelity, expanding the range of available operator classes, developing adaptive operator-library construction strategies and extending the framework to noisy experimental data and higher-dimensional scientific systems. These directions may further broaden the applicability of symbolic discovery to complex physical processes governed by non-local and history-dependent mechanisms.

\bibliographystyle{plain}
\bibliography{references}

\appendix

\section{Simulation Details} \label{app:simulation_details}

\subsection{Fractional Diffusion} \label{app:fractional_diffusion_simulation}

The fractional diffusion dataset was generated for the one-dimensional periodic problem
\begin{equation}
    u_t(x,t)+\kappa(-\Delta)^s u(x,t)=0,
\end{equation}
on $x\in[0,L)$ with $L=2\pi$. The parameters $\kappa=1.0$ and $s=0.7$ were fixed for all simulations. Only the initial condition $u(x,0)=u_0(x)$ was varied. The spatial domain was discretized using $N_x=256$ equidistant grid points, and the solution was stored at $N_t=51$ time levels on $t\in[0,1]$. In total, $100$ simulations were generated.

The fractional Laplacian was evaluated using a Fourier pseudospectral representation. For
\begin{equation}
    u(x,t)=\sum_k \hat{u}_k(t)e^{ikx},
\end{equation}
the fractional Laplacian satisfies
\begin{equation}
    \widehat{(-\Delta)^s u}_k=|k|^{2s}\hat{u}_k.
\end{equation}
Thus each Fourier mode evolves independently as
\begin{equation}
    \hat{u}_k(t)=\hat{u}_k(0)\exp\left(-\kappa |k|^{2s}t\right).
\end{equation}
This exact spectral evolution was applied at each stored time level, so no time-stepping scheme was required. The only numerical approximation is the finite Fourier truncation due to the spatial discretization.

The initial conditions were sampled as random mixtures of periodic Gaussian functions,
\begin{align}
    u_0(x)
    &=
    \sum_{j=1}^{N_g} a_j
    \exp\left(
        -\frac{d_{\mathrm{per}}(x,c_j)^2}{2\sigma_j^2}
    \right), \\
    d_{\mathrm{per}}(x,c_j)
    &=
    \min\left(|x-c_j|,L-|x-c_j|\right).
\end{align}
For each simulation, $N_g$ was sampled uniformly from the integers $2,\ldots,6$, with amplitudes $a_j\sim\mathcal{N}(0,1)$, centers $c_j\sim\mathcal{U}(0,L)$, and widths $\sigma_j\sim\mathcal{U}(0.04L,0.15L)$. After construction, the spatial mean of $u_0$ was subtracted to remove the undamped zero Fourier mode, and the result was normalized by its maximum absolute value.

\subsection{Euler--Poisson System} \label{app:euler_poisson_simulation}

The Euler--Poisson dataset was generated for the one-dimensional periodic system
\begin{equation}
    \rho_t + (\rho u)_x = 0, \qquad
    u_t + u\,u_x = -\phi_x, \qquad
    -\phi_{xx}=\rho-1,
\end{equation}
on $x\in[0,L)$ with $L=2\pi$. The spatial domain was discretized using $N_x=256$ equidistant grid points, matching the grid used for the fractional diffusion dataset. The solution was stored at $N_t=51$ time levels on $t\in[0,0.5]$. In total, $100$ simulations were generated.

The Poisson equation was solved using a Fourier pseudospectral representation. Writing the density perturbation as
\begin{equation}
    r(x,t)=\rho(x,t)-1
    =
    \sum_k \widehat{r}_k(t)e^{ikx},
\end{equation}
the periodic Poisson equation gives
\begin{equation}
    \widehat{\phi}_k(t)=\frac{\widehat{r}_k(t)}{k^2}, \qquad k\neq 0.
\end{equation}
The zero Fourier mode of $\phi$ was set to zero. The force was then computed spectrally as
\begin{equation}
    \widehat{(-\phi_x)}_k(t)
    =
    -ik\widehat{\phi}_k(t)
    =
    -\frac{i}{k}\widehat{r}_k(t),
    \qquad k\neq 0.
\end{equation}
The compatibility condition for the periodic Poisson problem was enforced by constructing the density with spatial mean equal to one.

The hyperbolic part of the system was advanced in time using a fourth-order Runge--Kutta method applied to
\begin{equation}
    \rho_t = -(\rho u)_x,
    \qquad
    u_t = -u u_x - \phi_x,
\end{equation}
where spatial derivatives were evaluated pseudospectrally. The internal time step was $\Delta t=10^{-3}$, while only the prescribed $N_t=51$ output time levels were stored.

The initial density and velocity were generated from independent random mixtures of periodic Gaussian functions using the same sampling procedure as in Appendix~\ref{app:fractional_diffusion_simulation}. For each simulation, a zero-mean random field $r_0(x)$ was generated and normalized by its maximum absolute value. The initial density was then defined as
\begin{equation}
    \rho_0(x)
    =
    1 + \varepsilon_\rho r_0(x),
    \qquad
    \varepsilon_\rho=0.4,
\end{equation}
which enforces $\langle \rho_0\rangle=1$ and gives an initial density in the approximate range $[0.6,1.4]$. The initial velocity was generated independently as
\begin{equation}
    u_0(x)
    =
    \varepsilon_u v_0(x),
    \qquad
    \varepsilon_u=0.3,
\end{equation}
where $v_0(x)$ is another normalized zero-mean periodic Gaussian-mixture field. These moderate-amplitude smooth initial conditions were used to make both the nonlinear convective term and the non-local Poisson force visible in the data while preserving a smooth solution regime over the simulated time interval.

For the symbolic discovery experiments, the input fields were taken to be the density perturbation $r=\rho-1$ and the velocity $u$. In these variables, the system can be written as
\begin{equation}
    r_t = -\partial_x((1+r)u),
    \qquad
    u_t = -u u_x - \phi_x,
    \qquad
    -\phi_{xx}=r.
\end{equation}
This formulation exposes the actual source of the Poisson equation directly to the neural-operator surrogate used for the non-local force term.

\subsection{Hereditary Viscoelasticity} \label{app:hereditary_viscoelasticity_simulation}

The hereditary viscoelasticity dataset was generated for the temporal constitutive law
\begin{equation}
    \sigma(t)=E\,\varepsilon(t)+\int_0^t A e^{-(t-s)/\tau}\,\varepsilon(s)\,ds,
\end{equation}
on $t\in[0,T]$ with $T=1$. The parameters were fixed to $E=1.0$, $A=0.5$, and $\tau=0.1$ for all simulations. The time interval was discretized using $N_t=256$ equidistant time points, and $100$ simulations were generated.

To evaluate the hereditary integral, we introduced the memory variable
\begin{equation}
    q(t)=\int_0^t e^{-(t-s)/\tau}\varepsilon(s)\,ds,
    \qquad
    \sigma(t)=E\varepsilon(t)+Aq(t).
\end{equation}
Equivalently, $q(t)$ satisfies
\begin{equation}
    \dot q(t)=\varepsilon(t)-\frac{1}{\tau}q(t),
    \qquad q(0)=0.
\end{equation}
On the uniform grid $t_n=n\Delta t$, the memory was advanced using the exact exponential update obtained from a piecewise-linear approximation of $\varepsilon(t)$ over each time interval:
\begin{equation}
    q_n
    =
    \alpha q_{n-1}
    +
    w_0\varepsilon_{n-1}
    +
    w_1\varepsilon_n,
    \qquad
    \alpha=\exp(-\Delta t/\tau),
\end{equation}
with
\begin{equation}
    w_0=
    \frac{\tau^2}{\Delta t}(1-\alpha)-\tau\alpha,
    \qquad
    w_1=
    \tau-\frac{\tau^2}{\Delta t}(1-\alpha).
\end{equation}
The stress was then computed as $\sigma_n=E\varepsilon_n+Aq_n$.

The strain histories were sampled as random mixtures of temporal Gaussian functions,
\begin{equation}
    \varepsilon(t)
    =
    \sum_{j=1}^{N_g}
    a_j
    \exp\left(
        -\frac{(t-c_j)^2}{2\sigma_j^2}
    \right),
\end{equation}
where $N_g$ was sampled uniformly from $20,\ldots,30$, $a_j\sim\mathcal{N}(0,1)$, $c_j\sim\mathcal{U}(0,T)$, and $\sigma_j\sim\mathcal{U}(0.03T,0.15T)$. Each strain history was centered by subtracting its temporal mean and normalized by its maximum absolute value.

The dataset stores $\varepsilon(t)$, the memory response $q(t)$, and the stress $\sigma(t)$. In the neural-operator part of the discovery pipeline, the relevant temporal non-local map is $\varepsilon(\cdot)\mapsto q(\cdot)$, while the full symbolic target is $\varepsilon(\cdot)\mapsto\sigma(\cdot)$. Thus, the neural operator represents the hereditary memory operation, and the symbolic model is expected to recover the algebraic combination $\sigma(t)=E\varepsilon(t)+Aq(t)$.

\end{document}